%% file: main.tex
\documentclass{article}

    \PassOptionsToPackage{numbers, compress}{natbib}
 \usepackage[preprint]{neurips_2025}

\usepackage[utf8]{inputenc} 
\usepackage[T1]{fontenc}    
\usepackage{url}            
\usepackage{booktabs}       
\usepackage{amsfonts}       
\usepackage{nicefrac}       
\usepackage{microtype}      
\usepackage{xcolor}         

\usepackage{pifont}
\usepackage{url}            
\usepackage{booktabs}       
\usepackage{amsfonts}       
\usepackage{nicefrac}       
\usepackage{microtype}      
\usepackage{xcolor}         
\usepackage{listings}
\usepackage{booktabs}
\usepackage{tabularx}
\usepackage{array}
\usepackage{amsmath}
\usepackage{amssymb}
\usepackage{adjustbox}
\usepackage{tcolorbox}
\usepackage{multirow}
\usepackage{makecell} 
\usepackage{tabularx}
\usepackage{graphicx}
\usepackage{subcaption}

\definecolor{darkblue}{rgb}{0, 0.12, 0.55}
\definecolor{darkgreen}{rgb}{0, 0.55, 0.12}
\definecolor{darkred}{rgb}{0.6,0,0}
\definecolor{darkgreen}{rgb}{0,0.6,0}
\definecolor{Gray}{gray}{0.9}
\usepackage[breaklinks=true,
            colorlinks,
            linkcolor = darkred,
            urlcolor  = magenta, 
            citecolor = teal,
            bookmarks = false]{hyperref}

\usepackage{booktabs}
\usepackage{multicol}
\usepackage{array}
\usepackage{xcolor}
\usepackage{colortbl}

\usepackage[accsupp]{axessibility}  
\usepackage{xcolor}
\usepackage{pifont}
\newcommand{\cmark}{\textcolor{green!60!black}{\ding{51}}} 
\newcommand{\xmark}{\textcolor{red!80!black}{\ding{55}}}  

\usepackage{hyperref}

\usepackage{orcidlink}

\begin{document}

\title{ManiTwin: Scaling Data-Generation-Ready Digital Object Dataset to 100K}

\author{%
Kaixuan Wang$^{1*}$ \And
Tianxing Chen$^{1,2*}$ \And
Jiawei Liu$^{10*}$ \And
Honghao Su$^{10*}$ \And
Shaolong Zhu$^{2*}$ \And
Minxuan Wang$^{10}$ \And
Zixuan Li$^{10}$ \And
Yue Chen$^{8}$ \And
Huan-ang Gao$^{9}$ \And
Yusen Qin$^{7}$ \And
Jiawei Wang$^{3,6}$ \And
Qixuan Zhang$^{3,5}$ \And
Lan Xu$^{5}$ \And
Jingyi Yu$^{5}$ \And
Yao Mu$^{4,\dagger}$ \And
Ping Luo$^{1,\dagger}$\AND
\textmd{$^1$The University of Hong Kong }\And
\textmd{$^2$Xspark AI} \And
\textmd{$^3$Deemos Tech} \And
\textmd{$^4$Shanghai Jiao Tong University} \And
\textmd{$^5$ShanghaiTech University} \And
\textmd{$^6$University of California, San Diego} \And
\textmd{$^7$D-Robotics} \And
\textmd{$^8$Peking University} \And
\textmd{$^9$Tsinghua University} \And
\textmd{$^{10}$Shenzhen University} \AND
\small{\rm{$^*$Equal Contribution  $^\dagger$Corresponding Authors}} \AND
\href{https://manitwin.github.io/}{https://manitwin.github.io}
}

%



\maketitle
\vspace{-5pt}
\begin{figure}[h]
    \centering
    \includegraphics[width=0.95\linewidth]{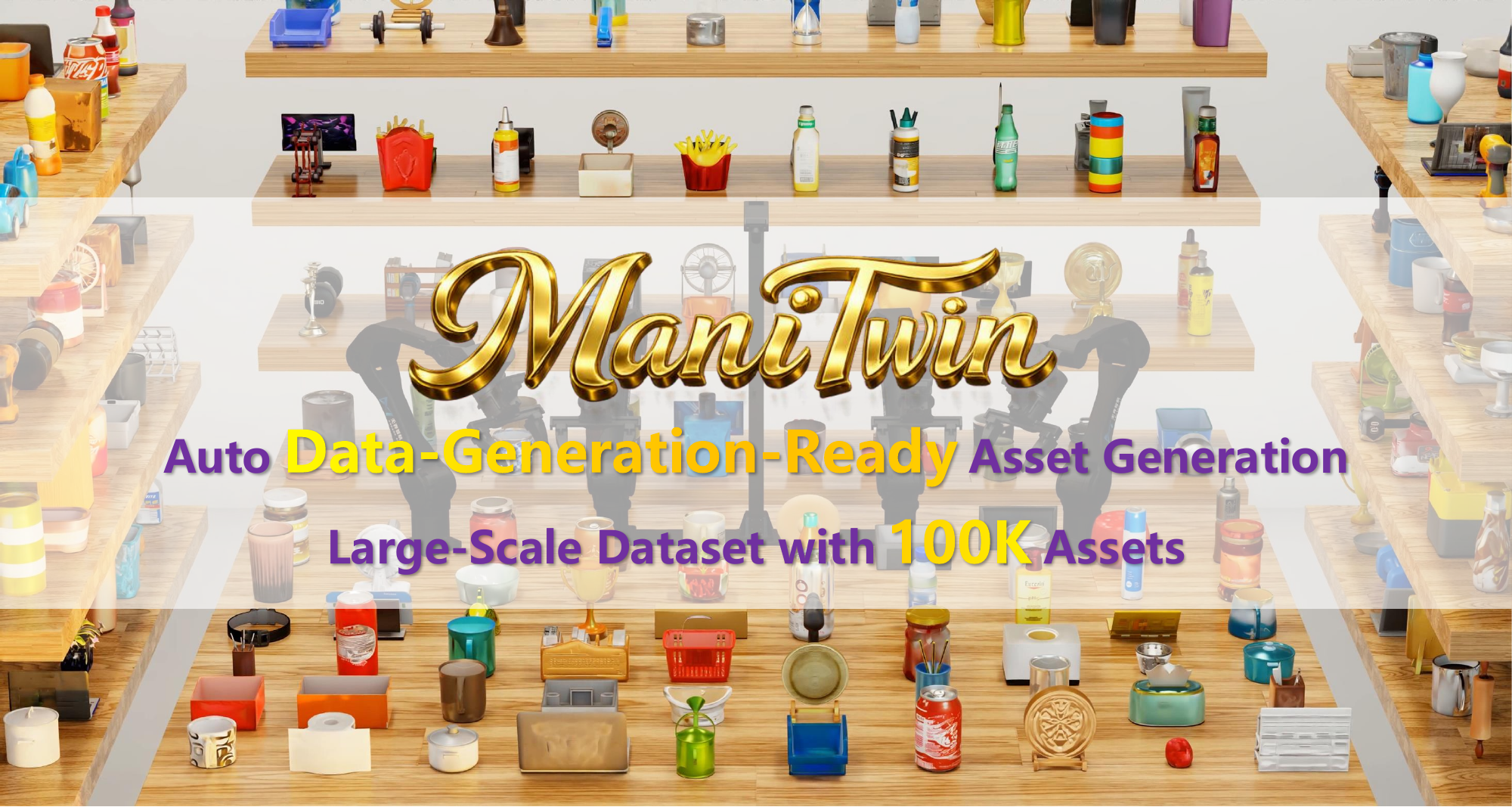}
    \caption{\textbf{ManiTwin.} An automated pipeline for synthesizing data-generation-ready assets and rich annotations, and \textbf{ManiTwin-100K}, a large-scale dataset of 100K such assets with diverse, high-quality annotations.}
    \label{fig:image_to_3d}
\end{figure}

\begin{abstract}
Learning in simulation provides a useful foundation for scaling robotic manipulation capabilities. However, this paradigm often suffers from a lack of data-generation-ready digital assets, in both scale and diversity. In this work, we present ManiTwin, an automated and efficient pipeline for generating data-generation-ready digital object twins. Our pipeline transforms a single image into simulation-ready and semantically annotated 3D asset, enabling large-scale robotic manipulation data generation. Using this pipeline, we construct ManiTwin-100K, a dataset containing 100K high-quality annotated 3D assets. Each asset is equipped with physical properties, language descriptions, functional annotations, and verified manipulation proposals. Experiments demonstrate that ManiTwin provides an efficient asset synthesis and annotation workflow, and that ManiTwin-100K offers high-quality and diverse assets for manipulation data generation, random scene synthesis, and VQA data generation, establishing a strong foundation for scalable simulation data synthesis and policy learning.
\end{abstract}

\input{section/introduction}

\input{section/related_work}

\input{section/method}

\input{section/application_scenarios}

\input{section/experiment}

\input{section/conclusion}

\section*{Acknowledgements}
We acknowledge Weiyang Jin for his valuable discussion during this project.

\newpage
\input{section/appendix}


%
%
\bibliographystyle{splncs04}
\bibliography{main}
\end{document}

%% file: section/introduction.tex
\section{Introduction}

Robotic manipulation learning in simulation critically depends on large-scale, high-quality object assets that not only represent geometric diversity, but also encode how objects can be physically interacted with. While 3D asset repositories have grown rapidly, most existing datasets are target geometric or visual understanding rather than manipulation-centric robotics. As a result, current robot learning researchers often face a fundamental mismatch between the assets used for training and the physical requirements of real-world manipulation.

Existing datasets expose a persistent gap between scale, semantic richness, and physical usability. Large-scale geometry datasets provide millions of meshes but lack physical parameters and interaction semantics, requiring extensive manual curation for robotics applications. Conversely, robotics-oriented datasets introduce articulation models or functional labels but remain limited in scale and do not systematically verify physical validity. What is missing is a large-scale dataset that simultaneously provides manipulation-centric objects, rich functional and grasp annotations aligned with language descriptions, and assets verified to be simulation-ready and collision-free.

To address this gap, we introduce \textbf{ManiTwin}, an automated pipeline for generating data-generation-ready digital object twins at scale. Starting from a single input image, ManiTwin synthesizes simulation-ready 3D assets, then employs vision-language models (VLMs) to annotate physical properties, functional points, and language descriptions. Candidate interaction points are sampled via farthest point sampling and filtered through VLM reasoning. A learning-based grasp generator produces diverse grasp proposals, which are screened by proximity to selected interaction points and validated through physics simulation. The resulting assets are high-fidelity, collision-ready, and equipped with rich manipulation annotations.

Our main contributions are:
\begin{itemize}
    \item We propose an automated pipeline that transforms a single input image into high-fidelity digital object twins with rich language, manipulation, and functional annotations, while ensuring that the generated assets are simulation-ready and collision-ready.
    \item Building upon this pipeline, we construct \textbf{ManiTwin-100K}, a large-scale dataset containing 100K semantically annotated digital assets, which supports a wide range of applications including simulation-based manipulation data synthesis, scene layout generation, and VQA data synthesis.
    \item We conduct extensive experiments to validate the quality and diversity of ManiTwin, demonstrating that ManiTwin provides a strong foundation for large-scale robotic manipulation data generation.
\end{itemize}

%% file: section/related_work.tex
\section{Related Works}

\subsection{Digital Object Twin Asset Dataset}

\input{table/compare_table}

Large-scale geometry datasets such as ShapeNet~\cite{chang2015shapenet}, ModelNet~\cite{fang2024modelnet}, Objaverse~\cite{deitke2023objaverse}, and Objaverse-XL~\cite{deitke2023objaverse-xl} provide extensive mesh collections with broad category coverage and language captions. However, these datasets are geometry-centric, containing many decorative or static objects unsuitable for manipulation. They lack physical parameters, articulation structures, or collision-validated assets, limiting their direct use in physics-based simulation.

To support interactive tasks, datasets such as PartNet-Mobility~\cite{xiang2020sapien}, GAPartNet~\cite{geng2023gapartnet}, and PhysXNet~\cite{cao2025physx} introduce articulated structures, part-level semantics, and affordance labels. However, these datasets remain limited in scale, often require post-processing for simulation deployment, and generally lack language descriptions. PASG~\cite{zhu2025pasg} proposes automatic annotation but without physics verification or a large-scale curated dataset.

High-quality collections such as YCB~\cite{calli2015ycb} emphasize physical fidelity but cover only tens of objects. RoboTwin-OD~\cite{chen2025robotwin} provides simulation-ready assets with manipulation annotations but lacks functional annotations and remains limited in scale. ManiTwin-100K bridges this gap by providing 100K manipulation-centric digital twins with functional and grasp annotations, language descriptions, and simulation-verified physical validity. We compare serval datasets with ManiTwin-100K in Table.~\ref{tab:dataset-comparison}.

\subsection{Data Generation in Simulation}

Simulation-based data synthesis has become a widely adopted approach for scaling robotic learning across diverse tasks. A representative line of work focuses on trajectory and interaction data synthesis based on annotated object assets. RoboTwin 1.0 series~\cite{mu2025robotwin,mu2024robotwin,chen2025benchmarking} explores synthesizing manipulation trajectories by combining asset-level manipulation annotations with expert code, while RoboTwin 2.0~\cite{chen2025robotwin} further introduces a larger annotated asset collection, RoboTwin-OD, and supports cluttered scene layouts to synthesize large-scale data spanning 50 manipulation tasks. Related efforts, including RoboGen~\cite{wang2023robogen}, InternData-A1~\cite{tian2025interndata}, UniVTAC~\cite{chen2026univtac}, RMBench~\cite{chen2026rmbench} and HumanoidGen~\cite{jing2025humanoidgen}, follow a similar paradigm of generating manipulation data from annotated assets, with InternData-A1 reporting over 7K hours of synthesized manipulation data.

Beyond action trajectories, several works target language-conditioned data synthesis. RoboRefer~\cite{zhou2025roborefer}, RoboTracer~\cite{zhou2025robotracer}, and Vlaser~\cite{yang2025vlaser} synthesize large-scale VQA data in simulation to support vision-language-action learning. Despite their success, these approaches fundamentally rely on simulation-ready assets with manipulation annotations, which constrains scalability and diversity. ManiTwin addresses this limitation by providing an automated pipeline for asset generation, annotation, and verification, enabling large-scale synthesis of physically valid assets for simulation data generation.

%% file: table/compare_table.tex

\begin{table*}[t]
\centering
\footnotesize
\setlength{\tabcolsep}{5pt}
\renewcommand{\arraystretch}{1.15}

\caption{\textbf{Comparison of ManiTwin-100K with other asset datasets.} 
We compare typical asset datasets in terms of geometry, simulation and collision readiness, manipulation and semantic annotations, and dataset scale, where Sim-ready refers to whether the object can be directly load into mainstream robotic simulators, grasping annotation refers to pre-contact grasp poses or affordance, function annotation refers to part-wise function information, and language annotation allows semantic understanding for the objects.}
\label{tab:dataset-comparison}

\begin{adjustbox}{max width=\linewidth}
\begin{tabular}{lccccccc}
\toprule
\textbf{Asset Dataset} &
\textbf{Rigid} &
\textbf{Sim-Ready} &
\textbf{\makecell{Grasping\\ Annotation}} &
\textbf{\makecell{Function\\ Annotation}} &
\textbf{\makecell{Language\\ Annotation}} &
\textbf{\#Objs} \\
\midrule
Objaverse~\cite{deitke2023objaverse}          & \cmark & \xmark  & \xmark  & \xmark & \cmark & 818K \\
Objaverse-XL~\cite{deitke2023objaverse-xl}       & \cmark & \xmark  & \xmark  & \xmark & \cmark & 10M+ \\
Objaverse++~\cite{lin2025objaverse++}      
& \cmark & \xmark  & \xmark & \xmark & \xmark & 500K \\
PhysXNet~\cite{cao2025physx} & \cmark & \cmark  & \xmark  & \cmark & \cmark & 26K \\
PhysXNet-XL~\cite{cao2025physx}         & \cmark & \cmark  & \xmark  & \cmark & \cmark & 6M \\
PartNet~\cite{xiang2020sapien}             & \cmark & \cmark  & \xmark & \xmark & \cmark & 26K \\
PartNet-Mobility~\cite{xiang2020sapien}    & \cmark & \cmark  & \xmark  & \xmark & \cmark & 2K \\
YCB Object Set~\cite{calli2015ycb}      & \cmark & \xmark & \cmark  & \cmark & \cmark & 77 \\
ModelNet~\cite{fang2024modelnet}            & \cmark & \xmark  & \xmark & \xmark & \xmark & 127K \\
ShapeNet~\cite{chang2015shapenet}            & \cmark & \xmark  & \xmark  & \xmark & \xmark & 51K \\
RoboTwin-OD~\cite{chen2025robotwin}        & \cmark & \cmark  & \cmark& \xmark & \xmark & 731 \\
GAPartNet~\cite{geng2023gapartnet}           & \cmark & \cmark  & \xmark  & \cmark & \xmark & 4K \\
OmniObject3D~\cite{wu2023omniobject3d}        & \cmark & \xmark  & \xmark  & \xmark & \cmark & 6K \\
\midrule
ManiTwin-100K (ours)   & \cmark & \cmark & \cmark & \cmark & \cmark & 100K \\
\bottomrule
\end{tabular}
\end{adjustbox}
\end{table*}

%% file: section/method.tex
\section{Method}

\begin{figure*}[t]
    \centering
    \includegraphics[width=1.0\linewidth]{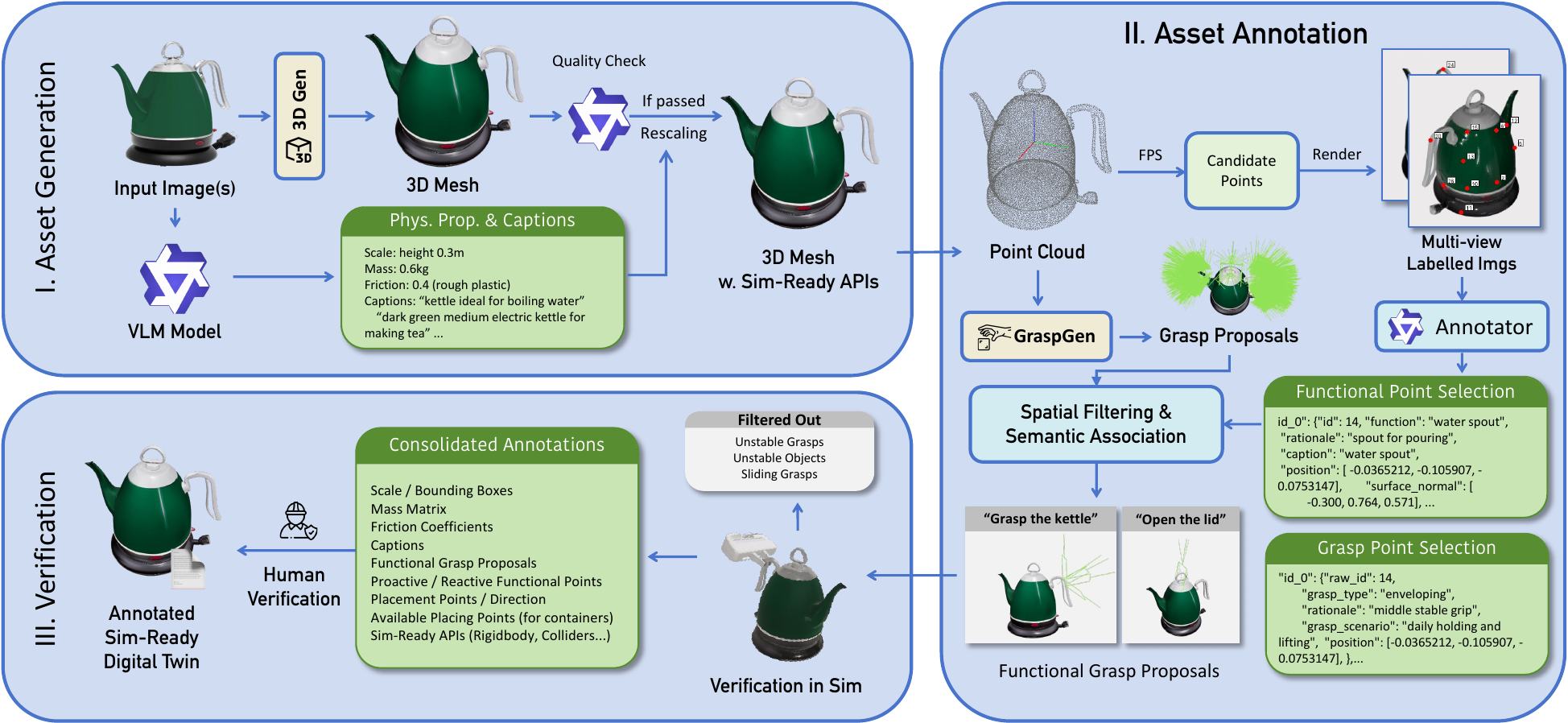}
    \caption{\textbf{ManiTwin Pipeline Overview.} Our pipeline consists of three stages: (I) \textit{Asset Generation} transforms input images into simulation-ready 3D meshes with VLM-estimated physical properties; (II) \textit{Asset Annotation} combines FPS-based candidate sampling, VLM-driven functional and grasp point selection, and learning-based grasp proposal generation; (III) \textit{Verification} validates annotations through physics simulation and human review, producing fully annotated digital twins ready for robotic manipulation research.}
    \label{fig:pipeline}
\end{figure*}

We present ManiTwin, an automated pipeline for generating data-generation-ready digital object twins at scale. As illustrated in Fig.~\ref{fig:pipeline}, our approach includes three stages: (I) Asset Generation, which transforms 2D images into simulation-ready 3D meshes; (II) Asset Annotation, which enriches assets with functional and manipulation semantics; and (III) Verification, which ensures physical validity through simulation and human review.

\subsection{Asset Generation}

The asset generation stage converts input images into physically-grounded 3D meshes suitable for robotic simulation.

\textbf{3D Generation.}
Given one or more input images depicting an object or a piece of text, we employ a state-of-the-art 3D generative model~\cite{zhang2024clay} to synthesize high-fidelity meshes. Input images undergo preprocessing including background removal and resolution normalization to isolate target objects. The generated meshes are converted to various formats with sim-ready APIs for compatibility with downstream simulation platforms. 

\textbf{Quality Verification.}
Not all generated assets are suitable for manipulation tasks, as the images and text are also generated. We employ a VLM-based quality gate that evaluates multi-view renderings against two criteria: (i) \textit{object singularity}---exactly one coherent object should be present; and (ii) \textit{visual quality}---the mesh should be free of severe artifacts such as fragmentation, texture corruption, or geometric implausibility. Assets failing these checks are filtered from the pipeline, removing approximately 10--15\% of generated content.

\textbf{Physical Property Estimation.}
For assets passing quality verification, we estimate physical properties essential for realistic simulation. A VLM analyzes eight uniformly-distributed renderings to infer: oriented bounding box (OBB) dimensions, estimated mass based on apparent material and volume, and surface friction coefficients derived from visual material cues. These estimates enable scale normalization to real-world dimensions via uniform scaling.

\textbf{Semantic Captioning.}
The VLM additionally generates rich semantic descriptions including object category, color, material, size, shape, and functional purpose. These language annotations support downstream applications such as language-conditioned manipulation and VQA data synthesis.

\subsection{Asset Annotation}

The annotation stage enriches each asset with manipulation-relevant semantics through a combination of geometric sampling, VLM-based reasoning, and learning-based grasp synthesis.

\begin{figure}[t]
    \centering
    \includegraphics[width=1.0\linewidth]{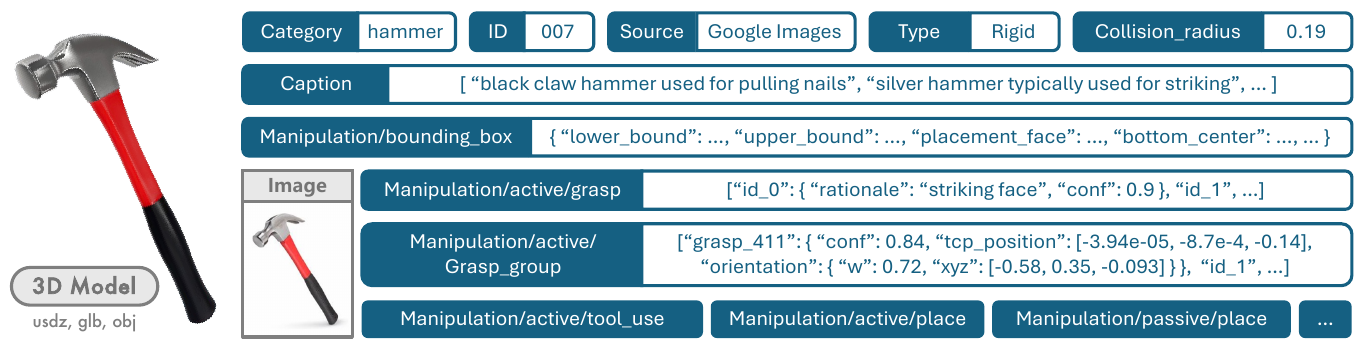}
    \caption{\textbf{Annotation Visualization.} ManiTwin provides functional points (e.g., spout, handle), grasp points with type labels, and simulation-verified 6-DoF grasp poses for each object.}
    \label{fig:object_annotation}
\end{figure}

\textbf{Candidate Point Sampling.}
To identify potential interaction regions, we sample a dense point cloud from the mesh surface and apply farthest point sampling (FPS) to select spatially-distributed candidates. FPS iteratively selects points maximizing minimum distance to the current set:
\begin{equation}
    p_{k+1} = \arg\max_{p \in \mathcal{P} \setminus \mathcal{S}} \min_{p_i \in \mathcal{S}} \|p - p_i\|_2,
\end{equation}
ensuring broad coverage across the object surface. These candidates are visualized as numbered markers on multi-view renderings for subsequent VLM annotation.

\textbf{VLM-Based Point Selection.}
We query a VLM with the labeled multi-view images to identify two types of semantically meaningful points:

\textit{Functional Points} correspond to regions with specific object functions---spouts for pouring, handles for gripping, blades for cutting, buttons for control. For each selected point, the VLM provides a function description, confidence score, and brief rationale.

\textit{Grasp Points} identify locations suitable for stable robotic grasping, considering proximity to center of gravity, surface geometry, and safety. Each grasp point includes a grasp type (parallel-jaw, pinch, power, three-finger, or enveloping) and intended use scenario - this allows different grasping poses to be conducted while doing different tasks.

\textbf{Grasp Proposal Generation.}
Building on VLM-selected points, we generate dense grasp proposals using GraspGen~\cite{murali2025graspgen}, a learning-based method that predicts stable grasp configurations from point cloud observations. Each proposal comprises a 6-DoF pose (position and quaternion orientation) with an associated confidence score.

\textbf{Spatial Filtering and Semantic Association.}
Raw grasp proposals are filtered by spatial proximity to VLM-selected points, retaining grasps aligned with identified affordances. Each filtered grasp inherits semantic labels from its nearest functional and/or grasp point, enabling task-oriented grasp selection (e.g., ``grasp the handle for pouring'').


\subsection{Verification}

The verification stage ensures that all annotations are physically valid and ready for deployment in simulation environments.

\textbf{Simulation Verification.}
Each grasp proposal undergoes physics-based validation using the SAPIEN simulator with PhysX. We execute a standardized grasp sequence: position the gripper, close fingers until contact, and verify stability. Successful grasps must maintain stable contact for multiple consecutive frames. We further test robustness through a slide resistance protocol, moving the grasped object along orthogonal directions and discarding grasps where object displacement exceeds a threshold. Only grasps passing both stability and slide tests are retained.

\textbf{Human Verification.}
While automated verification ensures physical validity, human annotators review a sampled subset to catch edge cases. Annotators assess mesh quality, physical plausibility of estimated properties, semantic correctness of annotations, and simulation results. Feedback is aggregated to iteratively refine VLM prompts and filtering thresholds.

\textbf{Consolidated Output.}
The final output for each object comprises: (i) simulation-ready 3D mesh with PBR materials; (ii) physical properties (dimensions, mass, friction); (iii) language annotations; (iv) functional and grasp point annotations with semantic labels; (v) simulation-verified 6-DoF grasp poses; and (vi) placement annotations for scene generation. Representative annotations are visualized in Fig.~\ref{fig:object_annotation}.

\subsection{ManiTwin-100K Dataset}

Using the pipeline above, we construct ManiTwin-100K, a large-scale dataset of 100K data-generation-ready digital twins. Unlike existing 3D datasets that prioritize geometric diversity or visual fidelity alone, ManiTwin-100K is specifically designed to address the fundamental requirements of robotic manipulation research: simulation-ready assets with rich manipulation semantics and verified physical validity. ManiTwin-100K bridges this gap by providing \textit{both} scale and manipulation-centric annotations within a unified dataset, enabling large-scale policy learning across diverse object categories and interaction types.

\textbf{Data Collection.}
Input images are curated from diverse sources: e-commerce product catalogs capturing real-world object appearances and text-to-image generations for underrepresented categories. 

\textbf{Statistics and Distribution of ManiTwin-100K Dataset.}
ManiTwin-100K spans manipulation-relevant categories including kitchen items (cups, mugs, utensils, containers, bottles), tools (hammers, screwdrivers, pliers, wrenches), electronics (phones, remotes, controllers), personal items (brushes, cosmetics, accessories), office supplies (staplers, tape dispensers, pens), and household objects (cleaning supplies, toys, food items). 

Each object in ManiTwin-100K includes: 2--4 functional points with semantic labels, 2--3 grasp points with grasp type annotations, 10--50 simulation-verified 6-DoF grasp poses, physical properties (OBB dimensions, mass, friction), and rich language descriptions (category, color, material, shape, function) as shown in Fig.~\ref{fig:object_annotation}.

%% file: section/application_scenarios.tex
\section{Applications}
\label{sec:applications}

The rich semantic annotations in ManiTwin enable a variety of downstream applications in robotics and 3D vision. In this section, we highlight four key application scenarios: manipulation data generation, scene layout generation, robotics VQA data synthesis, and 3D understanding tasks.

\subsection{Manipulation Data Generation}

One of the primary applications of ManiTwin is automated generation of large-scale manipulation training data. The combination of simulation-ready meshes, verified grasp poses, and functional point annotations enables fully automated data collection pipelines for learning generalizable manipulation skills.

\textbf{Pick and Place Skills Data Generation.}
Each object in ManiTwin comes with simulation-verified 6-DoF grasp poses and placement vector that are guaranteed to be collision-free and physically stable. By placing objects in simulation environments and executing these grasp poses, we can automatically generate successful grasp demonstrations at scale. This eliminates the need for costly human teleoperation or manual grasp labeling, enabling the collection of millions of grasp trajectories across diverse objects.

\textbf{Functional Manipulation.}
Beyond simple pick-and-place, ManiTwin's functional point annotations enable task-oriented manipulation data generation. For instance, we can generate trajectories for grasping a mug by its handle for pouring, picking up a knife by its handle for cutting, or holding a spray bottle by its trigger for spraying. These functionally-grounded demonstrations provide richer supervision signals for learning manipulation policies that understand object affordances.

\textbf{Automated Task Generation.}
Beyond data collection for predefined tasks, ManiTwin's rich semantic annotations enable \textit{automated task generation} at scale. Given a generated scene layout, we can programmatically compose diverse manipulation tasks by leveraging object-level annotations: functional points define what actions are possible (pour, cut, press, open), grasp points specify how to interact, and language descriptions provide natural task specifications. For example, given a scene with a kettle and a mug, the system can automatically generate tasks like ``pour water from the kettle into the mug'' by identifying the kettle's spout (functional point) and the mug's opening (placement target).

This capability significantly reduces the human effort required to scale task diversity. Prior works such as RoboTwin2.0~\cite{chen2025robotwin} and RoboCasa365~\cite{robocasa365} rely on manual task specification, limiting scalability. In contrast, ManiTwin's annotation-driven approach can generate hundreds of task variants per scene combination, covering pick-and-place, tool use, pouring, insertion, and other manipulation primitives. Diverse tasks translate to diverse skills, and training data spanning this task distribution can benefit general-purpose manipulation policy learning.

\begin{figure}[t]
    \centering
    \begin{subfigure}[t]{0.49\linewidth}
        \centering
        \includegraphics[width=\linewidth]{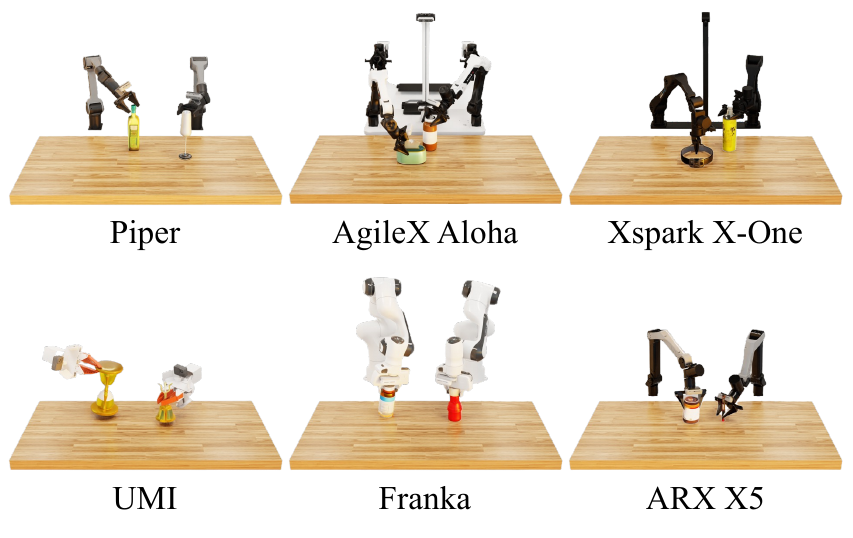}
        \caption{\textbf{Cross-Embodiment Manipulation Data Generation.} ManiTwin-100K enables automated data generation across different robotic platforms. We show manipulation trajectories generated for multiple end-effectors using the same underlying object annotations.}
        \label{fig:cross_embodiment}
    \end{subfigure}\hfill
    \begin{subfigure}[t]{0.49\linewidth}
        \centering
        \includegraphics[width=\linewidth]{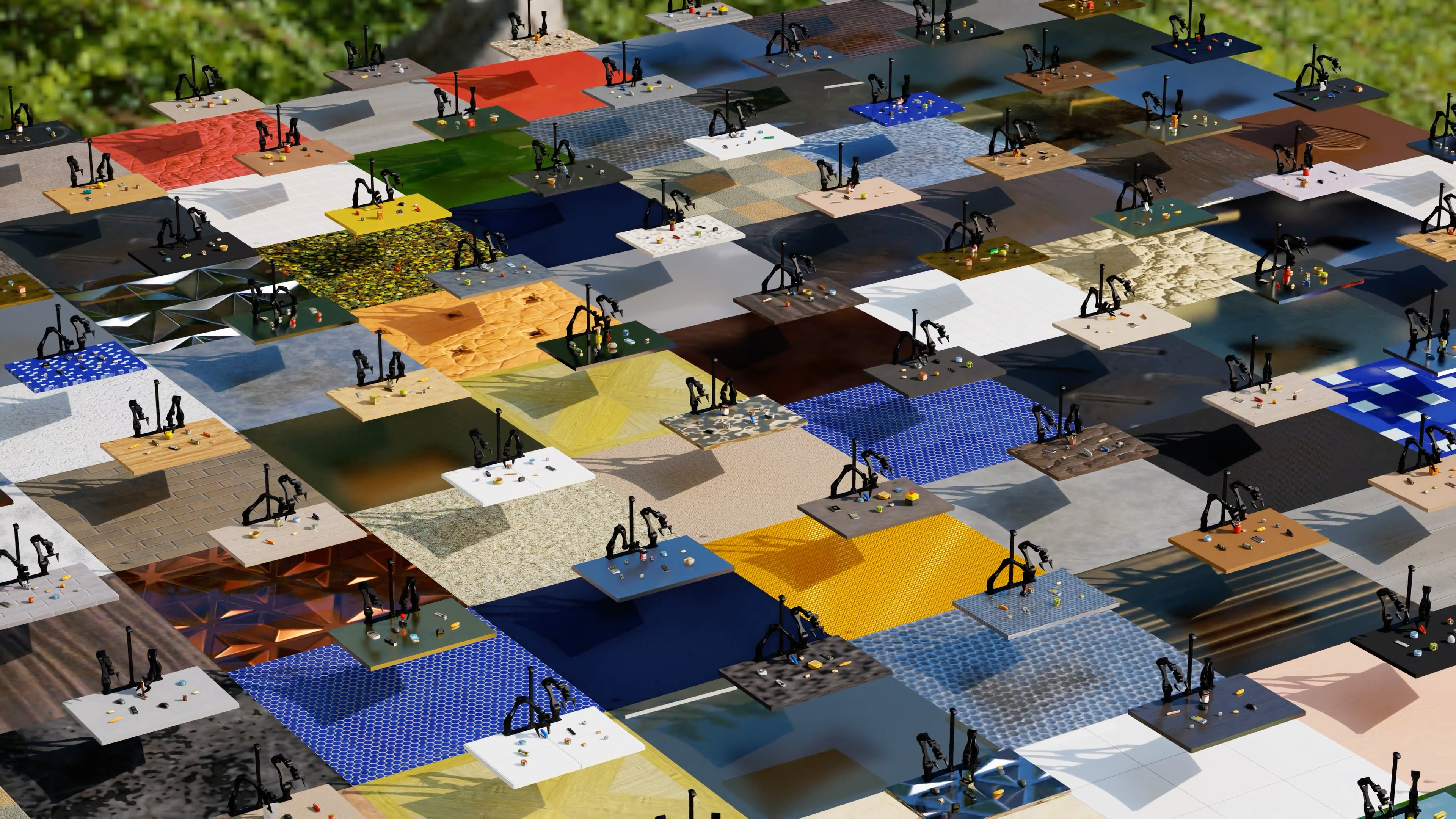}
        \caption{\textbf{Large Scale Grasping Data Generation.} ManiTwin-100K enables automated data generation across different robotic platforms. We show manipulation trajectories generated for multiple end-effectors using the same underlying object annotations.}
        \label{fig:grasp_datagen}
    \end{subfigure}
    \caption{\textbf{ManiTwin Data Generation.} (Left) Cross-embodiment manipulation trajectories across multiple end-effectors using shared object annotations. (Right) Grasping data generation.}
    \label{fig:datagen_combined}
\end{figure}

\textbf{Cross-Embodiment Data.}
While grasp poses in ManiTwin are initially generated and verified for the Franka Panda gripper, the underlying grasp points and functional annotations transfer across different end-effectors. As shown in Fig.~\ref{fig:cross_embodiment}, we leverage ManiTwin to generate manipulation data for multiple robotic platforms, including parallel-jaw grippers, dexterous hands, and custom end-effectors. This cross-embodiment data generation capability supports pretraining of generalizable manipulation policies that can transfer across different robot hardware.


\begin{figure}[t]
    \centering
    \includegraphics[width=1.0\linewidth]{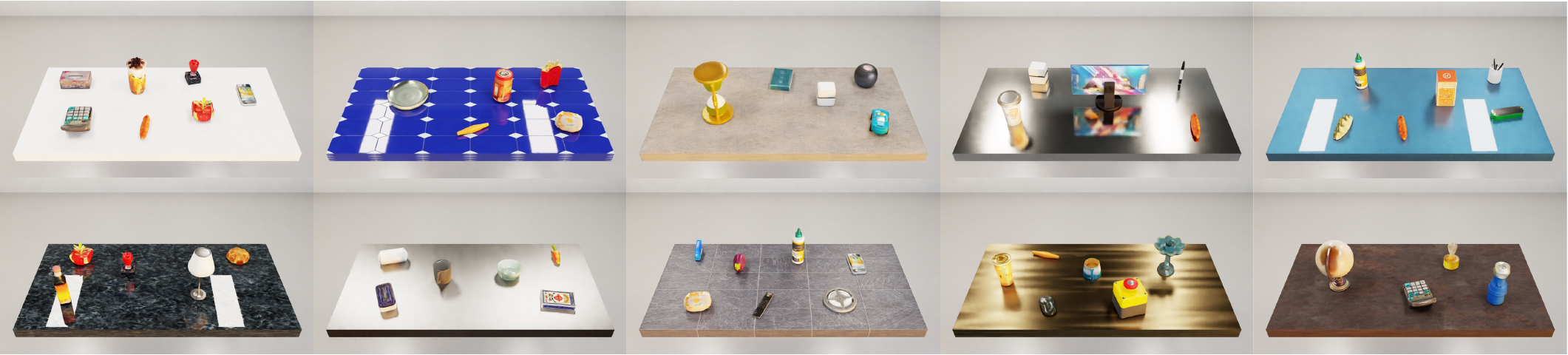}
    \caption{\textbf{Layout Generation.} Using placement and collision radius annotations, we generate diverse multi-object scene layouts that are collision-free and physically plausible.}
    \label{fig:clutter_layout}
\end{figure}

\subsection{Layout Generation}

With per-object placement annotations, including a placement position and orientation, we can deterministically place a single object on a supporting surface in simulation. To avoid overlaps when randomly generating multi-object scenes, we additionally provide a \texttt{collision\_radius} annotation, which defines the projected collision radius of each object on the placement plane. This allows us to sample object placements while preventing inter-object overlaps and collisions. Using these annotations, we can generate diverse random layouts. Fig.~\ref{fig:clutter_layout} visualizes randomly generated layouts of multiple objects on different table configurations. These layouts support manipulation policy data generation and evaluation with random tabletop distractors, and also serve as a basis for VQA data synthesis.

\subsection{VQA Data Generation}

\begin{figure}[h!]
    \centering
    \small
    \begin{tabularx}{\columnwidth}{@{} m{0.3\columnwidth} X @{}}
        \toprule
        \textbf{Visual Scene} & \textbf{VQA Training Pair} \\ \midrule

        \includegraphics[width=\linewidth]{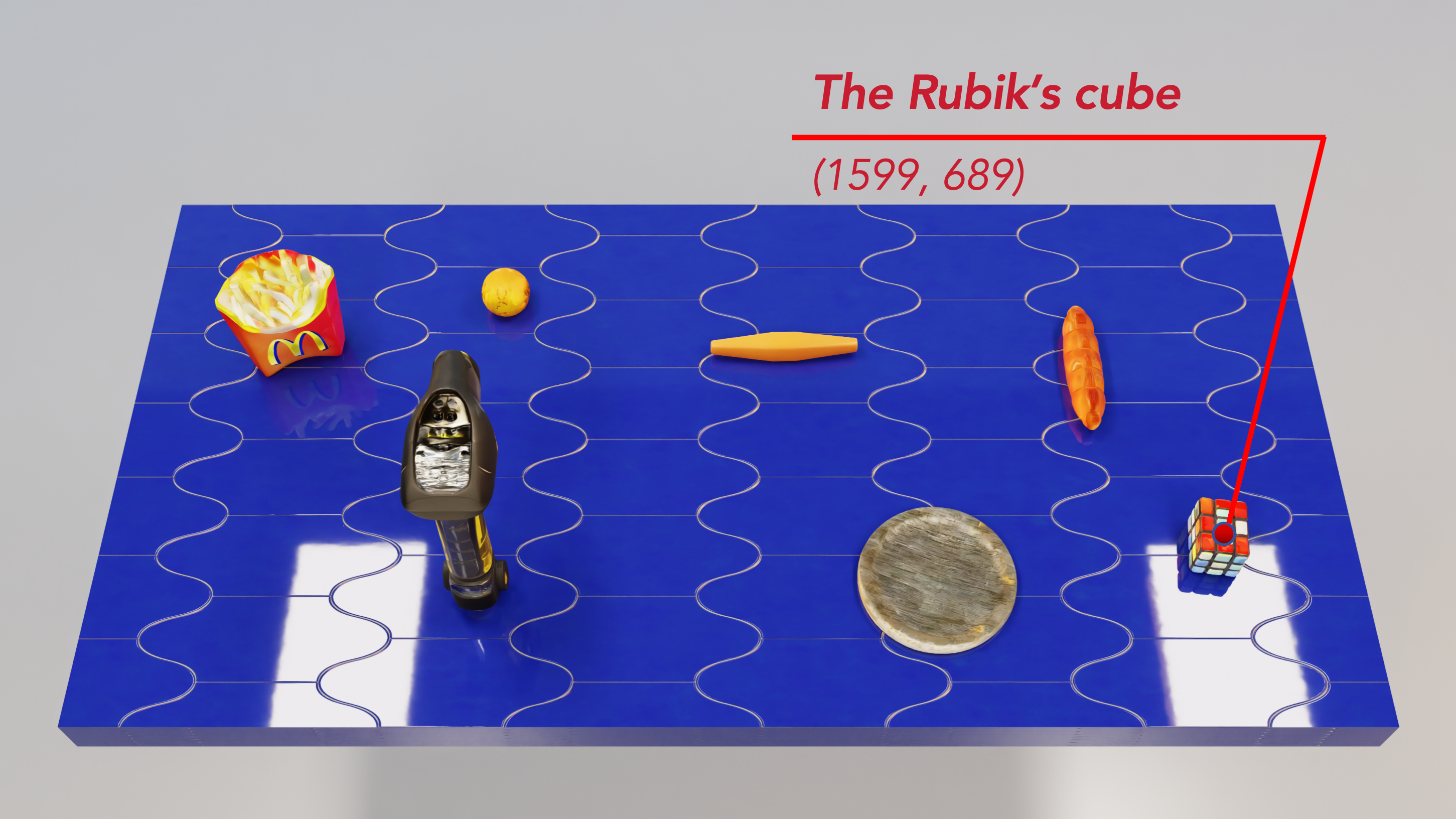} &
        \textbf{Q[Language Grounding]:} Locate the object with a grid of colored squares and describe its best grasp type. \par
        \textbf{A:} The Rubik's cube at the bottom right. Its flat, orthogonal faces are ideal for a stable parallel-jaw grasp. \\ \midrule

        \includegraphics[width=\linewidth]{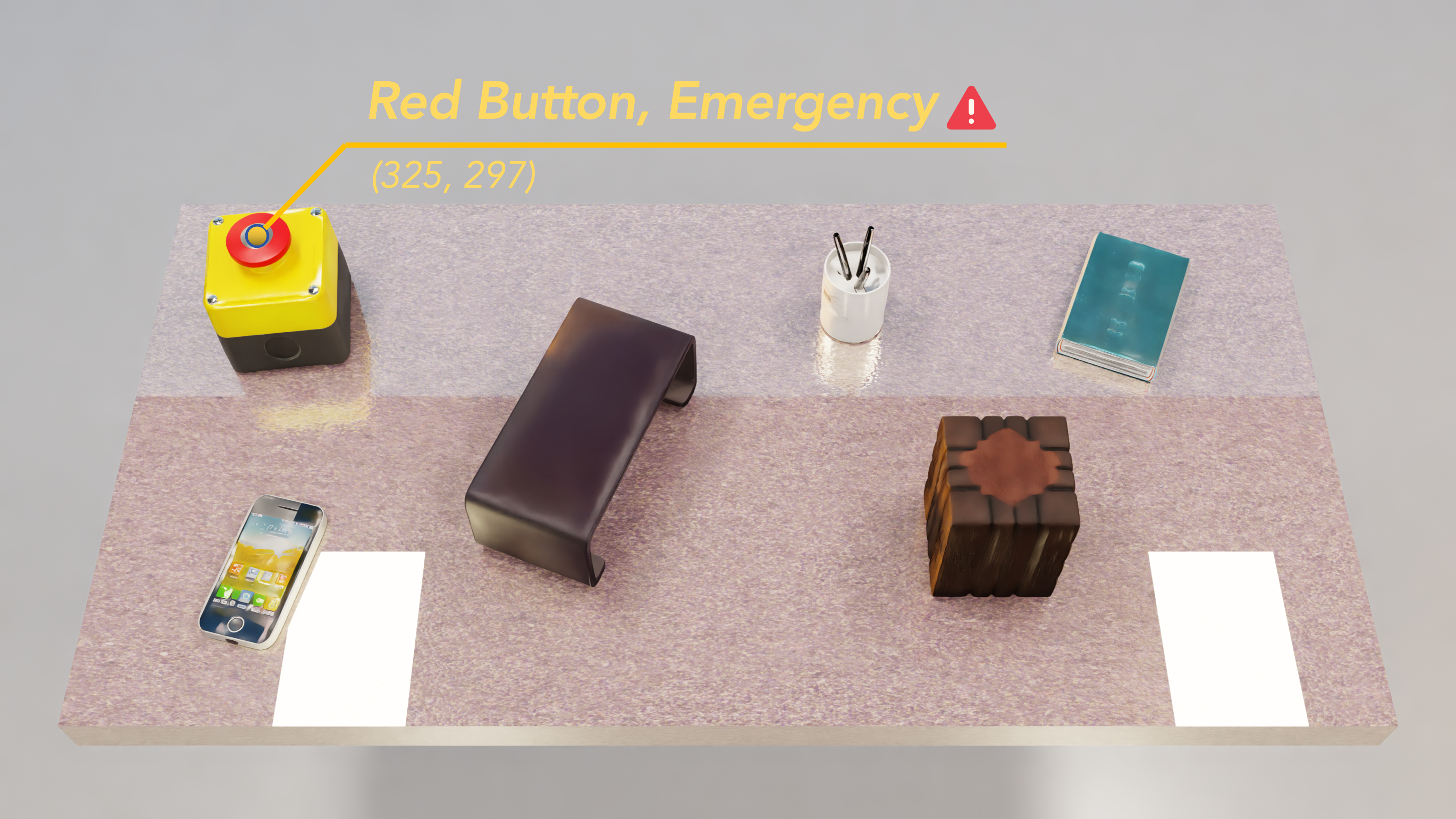} &
        \textbf{Q[Functional Planning]: } Where is the interaction point of the emergency safety device, and is it occluded by the riser? \par
        \textbf{A:} The red circular button on the top-left housing. It is not occluded by the central black riser and is accessible via a top-down approach. \\ \midrule

        \includegraphics[width=\linewidth]{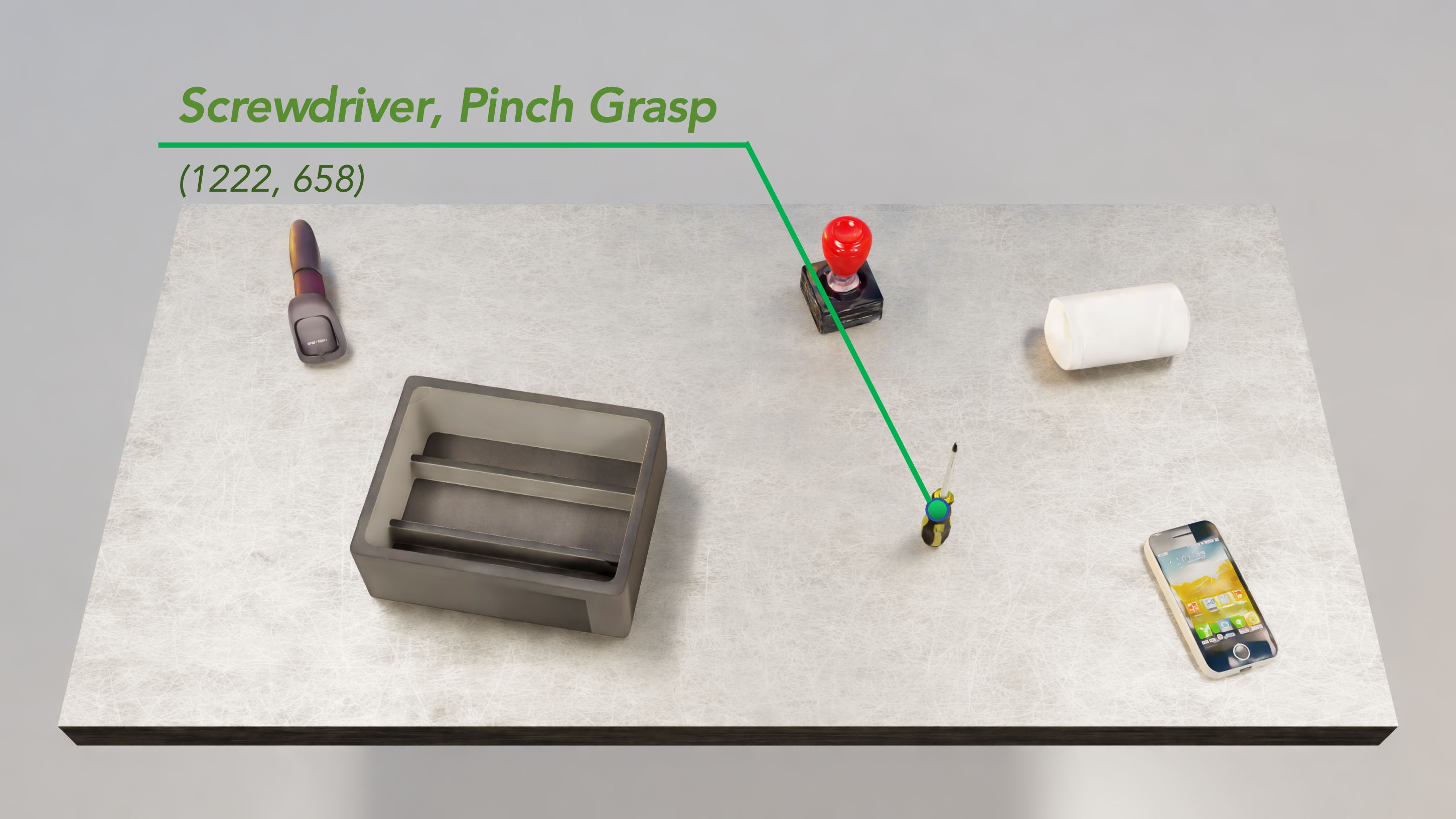} &
        \textbf{Q[Scene Understanding]: } Which tool requires a precision ``pinch'' grasp, and is there enough clearance to its right? \par
        \textbf{A:} The screwdriver (yellow/black handle). There is insufficient clearance to the right due to the proximity of the white cloth roll. \\ \midrule

        \includegraphics[width=\linewidth]{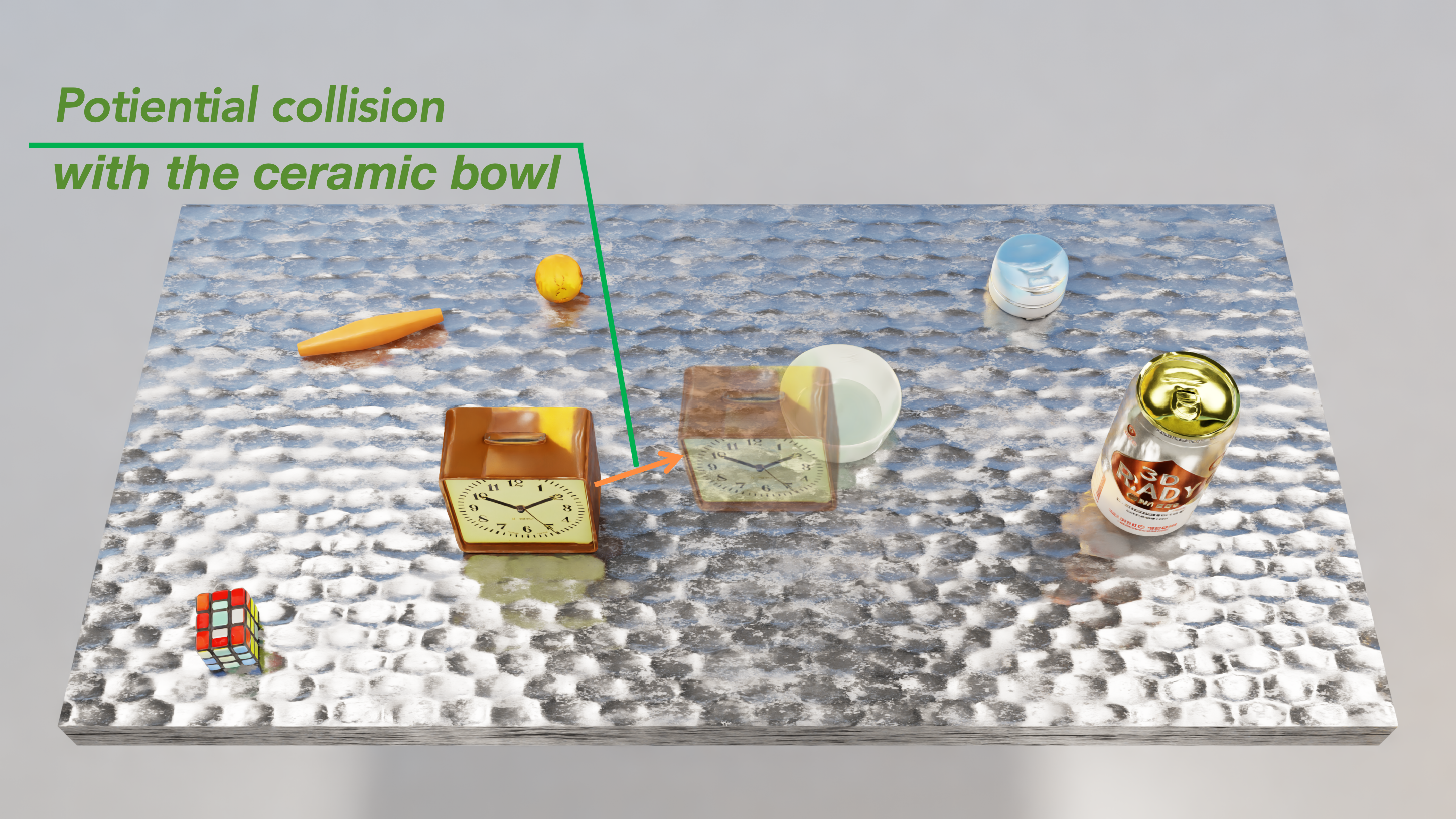} &
        \textbf{Q[Task Planning]: } If the robot slides the clock 10cm to the right then 5cm to the front, identify the risk of collision. \par
        \textbf{A:} There is a high collision risk with the white ceramic bowl, which is located directly in the clock's linear path. \\ \midrule

        \includegraphics[width=\linewidth]{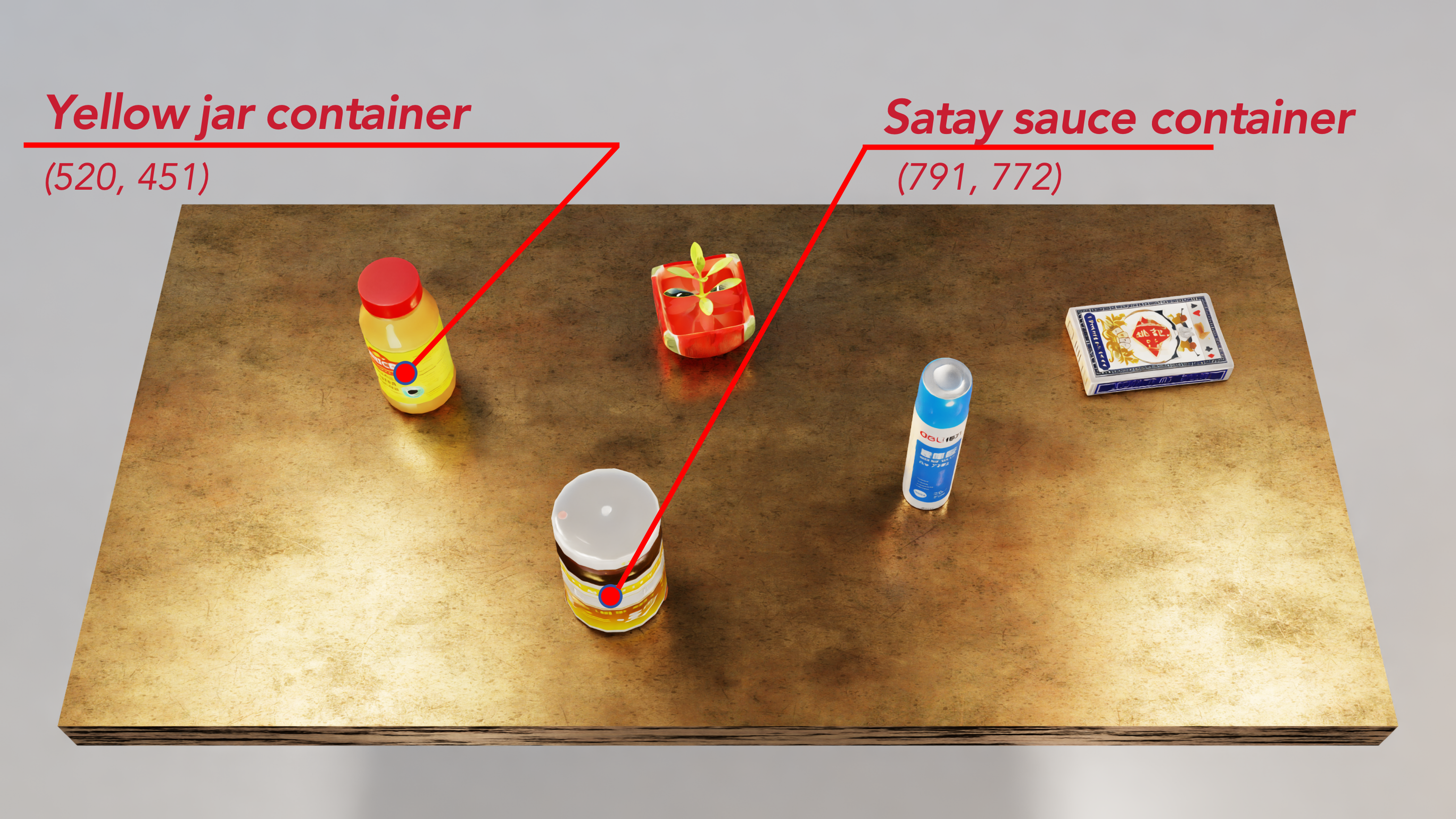} &
        \textbf{Q[Detection]: } Enumerate and identify all containers present on the brushed brass surface. \par
        \textbf{A:} There are two containers: one yellow vitamin jar (left), and one red/green seedling pot (right). \\

        \bottomrule
    \end{tabularx}
    \caption{\textbf{Robotics VQA Examples.} Each pair links manipulation-relevant questions to grounded scene understanding, covering language grounding, functional planning, scene understanding, task planning, and object detection.}
    \label{fig:vqa_grid}
\end{figure}

The rich semantic annotations in ManiTwin provide a foundation for generating Visual Question Answering (VQA) data specifically tailored for robotics applications. Unlike general-purpose VQA datasets, robotics-focused VQA requires understanding of manipulation affordances, physical properties, spatial reasoning, and action-relevant object attributes. We curate a VQA dataset across diverse tabletop environments with varying surface textures (metallic, wood, marble) and object densities to simulate realistic manipulation scenarios. Fig.~\ref{fig:vqa_grid} illustrates representative examples spanning five question categories.

The resulting VQA dataset can be used to train or fine-tune robotics-specific vision-language models (VLMs) that understand manipulation-relevant visual concepts and can provide actionable guidance for robotic task execution.

\subsection{3D Understanding Tasks}

As a large-scale 3D object dataset with dense semantic annotations, ManiTwin supports various 3D computer vision tasks beyond robotics applications.

\textbf{3D Part Segmentation.}
Combined with our layout generation capability, ManiTwin enables the creation of large-scale training data for 3D object segmentation in cluttered scenes. By rendering multi-object layouts with known object identities and positions, we can automatically generate ground truth instance segmentation masks for both 2D images and 3D point clouds. This supports training and evaluation of models for instance segmentation, semantic segmentation, and panoptic segmentation in realistic tabletop manipulation scenarios with varying levels of clutter and occlusion. 
  
\textbf{3D Object Understanding.}
ManiTwin provides multi-modal annotations including point clouds, meshes, multi-view images, and language descriptions for each object. This rich annotation enables research on 3D object classification, shape retrieval, cross-modal learning (e.g., text-to-3D, image-to-3D retrieval), and 3D captioning tasks.

\textbf{Affordance Prediction.}
The grasp points and functional points in ManiTwin can serve as ground truth for learning affordance prediction models. Given a 3D object representation (point cloud, mesh, or multi-view images), models can be trained to predict where and how the object can be grasped or manipulated, which is valuable for both robotics and human-object interaction understanding.


\subsection{Empowering Real-World Manipulation}

Recent works such as G3Flow~\cite{chen2025g3flow}, D(R,O) Grasp~\cite{wei2024mathcal} and CordViP~\cite{fu2025cordvip} suggest that high-quality 3D object assets, when combined with pose estimation tools like FoundationPose~\cite{wen2024foundationpose}, can bridge the gap between simulation and real-world manipulation. ManiTwin contributes to this direction by providing manipulation-ready 3D object twins with accurate geometry, physical properties, and semantic annotations. In practice, a robot can reconstruct or retrieve an object twin from a single image, estimate its 6D pose in the scene, and then leverage the associated semantic information, grasp proposals and functional points to plan and execute actions. Moreover, our scalable asset generation makes it feasible to cover long-tail object instances and diverse appearances, which is important for robust manipulation in open-world settings.

%% file: section/experiment.tex
\section{Experiments}

We conduct comprehensive experiments to evaluate the quality of ManiTwin assets and demonstrate their utility for downstream applications. We assess: (1) 3D generation quality through latent-based metrics; (2) annotation quality through automated verification and human evaluation; and (3) grasp quality and data generation statistics.

\subsection{3D Generation Quality}

We evaluate the quality of generated 3D assets using latent-based metrics that measure semantic alignment between generated 3D assets and their input conditions (images or text).

\textbf{Evaluation Metrics.}
We adopt latent-based metrics to assess both geometric and appearance quality. For geometry evaluation, CLIP(N-I) and CLIP(N-T) compute the CLIP similarity between rendered normal maps and the input image or text description, respectively, measuring how well the generated 3D geometry aligns with the input conditions. For appearance evaluation, CLIP(I-I) and CLIP(I-T) compute the CLIP similarity between rendered RGB images and the input image or text, assessing visual fidelity and semantic consistency. Additionally, ULIP-I and ULIP-T measure 3D-to-image and 3D-to-text alignment using the ULIP~\cite{xue2023ulip} latent space, providing a holistic assessment of how well the generated 3D asset matches the input semantically. For each metric, we compute the average score across all 30 rendered views.

\textbf{Results.}
We adopt the results from CLAY~\cite{zhang2024clay} in Table~\ref{tab:generation-quality} with the 3D generation quality metrics for both image-to-3D and text-to-3D generation. The results demonstrate that our pipeline produces semantically aligned 3D assets that faithfully preserve both the geometric structure and visual appearance of the input conditions. Image-to-3D generation achieves substantially higher scores across all metrics, reflecting the richer conditioning signal provided by input images compared to text descriptions.

\begin{table}[h]
\centering
\footnotesize

\begin{minipage}[t]{0.53\linewidth}
\centering
\setlength{\tabcolsep}{4pt}
\renewcommand{\arraystretch}{1.10}
\caption{\textbf{3D Generation Quality Evaluation.} Latent metrics for text-to-3D and image-to-3D. Geometry uses normal renders; appearance uses RGB renders.}
\label{tab:generation-quality}
\begin{tabular}{lcc}
\toprule
\textbf{Metric} & \textbf{Text-to-3D} & \textbf{Image-to-3D} \\
\midrule
ULIP & 0.1705 & 0.2140 \\
CLIP(N-I/T) & 0.1948 & 0.6848 \\
CLIP(I-I/T) & 0.2324 & 0.7769 \\
Time (s) & $\sim$45 & $\sim$45 \\
\bottomrule
\end{tabular}
\end{minipage}\hfill
\begin{minipage}[t]{0.45\linewidth}
\centering
\setlength{\tabcolsep}{6pt}
\renewcommand{\arraystretch}{1.10}
\caption{Data Generation Statistics. Summary of manipulation data generated with ManiTwin assets and annotations.}
\label{tab:datagen-stats}
\begin{tabular}{lc}
\toprule
\textbf{Statistic} & \textbf{Value} \\
\midrule
Total objects & 100K \\
Total verified grasps & 5M \\
Total grasp trajectories & 10M \\
Avg. trajectory length & 7.6s \\
\bottomrule
\end{tabular}
\end{minipage}

\end{table}







\subsection{Annotation Quality}

We evaluate the quality of ManiTwin annotations through both automated verification and human evaluation.

\textbf{Automated Verification.}
Our multi-stage pipeline incorporates automated quality checks at each stage. Table~\ref{tab:auto-verif} summarizes the key statistics. The 3D generation stage achieves a 69.67\% success rate, filtering out low-quality or failed generations. For grasp annotation, we generate an average of 81.63 grasp candidates per object through GraspGen and VLM-guided filtering. After physics-based simulation verification, 62.14 grasps per object are retained on average, corresponding to a 76.13\% verification success rate. This indicates that the majority of proposed grasps are physically valid and stable.

\textbf{Human Evaluation.}
We conduct a human study to evaluate the quality of VLM-generated annotations. Annotators assess a random sample of 500 objects across five dimensions to evaluate whether the 3D assets and annotations meet the requirements for manipulation data generation. Table~\ref{tab:human-eval} reports the results. Category classification achieves perfect accuracy, while language descriptions reach 99.6\% accuracy. Functional point labels and physical property estimation both achieve 92.2\% accuracy. Grasp point selection shows slightly lower accuracy at 84.8\%, reflecting the inherent difficulty of identifying optimal grasp locations purely from visual reasoning. Overall, the high accuracy across all annotation types validates the effectiveness of our VLM-based annotation approach.

\begin{table}[t]
\centering
\footnotesize
\setlength{\tabcolsep}{5pt} 
\caption{Automated and human evaluation of annotation quality.}
\label{tab:pipeline-human}

\begin{subtable}[t]{0.53\linewidth}
    \centering
    \caption{\textbf{Automated Verification.} Pass rate of the 3D generation and grasp verification in simulation, and the average grasp proposals per object before/after verification.}
    \label{tab:auto-verif}
    \begin{tabular}{lc}
        \toprule
        \textbf{Metric} & \textbf{Value} \\
        \midrule
        3D-gen succ. rate & 69.67\% \\
        Grasp verification succ. rate & 76.13\% \\ \midrule
        Avg. grasp candidates / obj. & 81.63 \\
        Avg. verified grasps / obj. & 62.14 \\
        \bottomrule
    \end{tabular}
\end{subtable}
\hfill
\begin{subtable}[t]{0.45\linewidth}
    \centering
    \caption{Human Evaluation of Annotation Quality. Percent of annotations judged correct on 500 sampled objects.}
    \label{tab:human-eval}
    \begin{tabular}{lc}
        \toprule
        \textbf{Annotation Type} & \textbf{Acc. (\%)} \\
        \midrule
        Category Classification & 100.0 \\
        Language Descriptions & 99.6 \\
        Functional Point Labels & 92.2 \\
        Phys. Prop. Estimation & 92.2 \\
        Grasp Point Selection & 84.8 \\
        \bottomrule
    \end{tabular}
\end{subtable}
\end{table}

\subsection{Analysis on ManiTwin-100K Diversity}

\begin{figure}[t]
    \centering
    \includegraphics[width=0.35\linewidth]{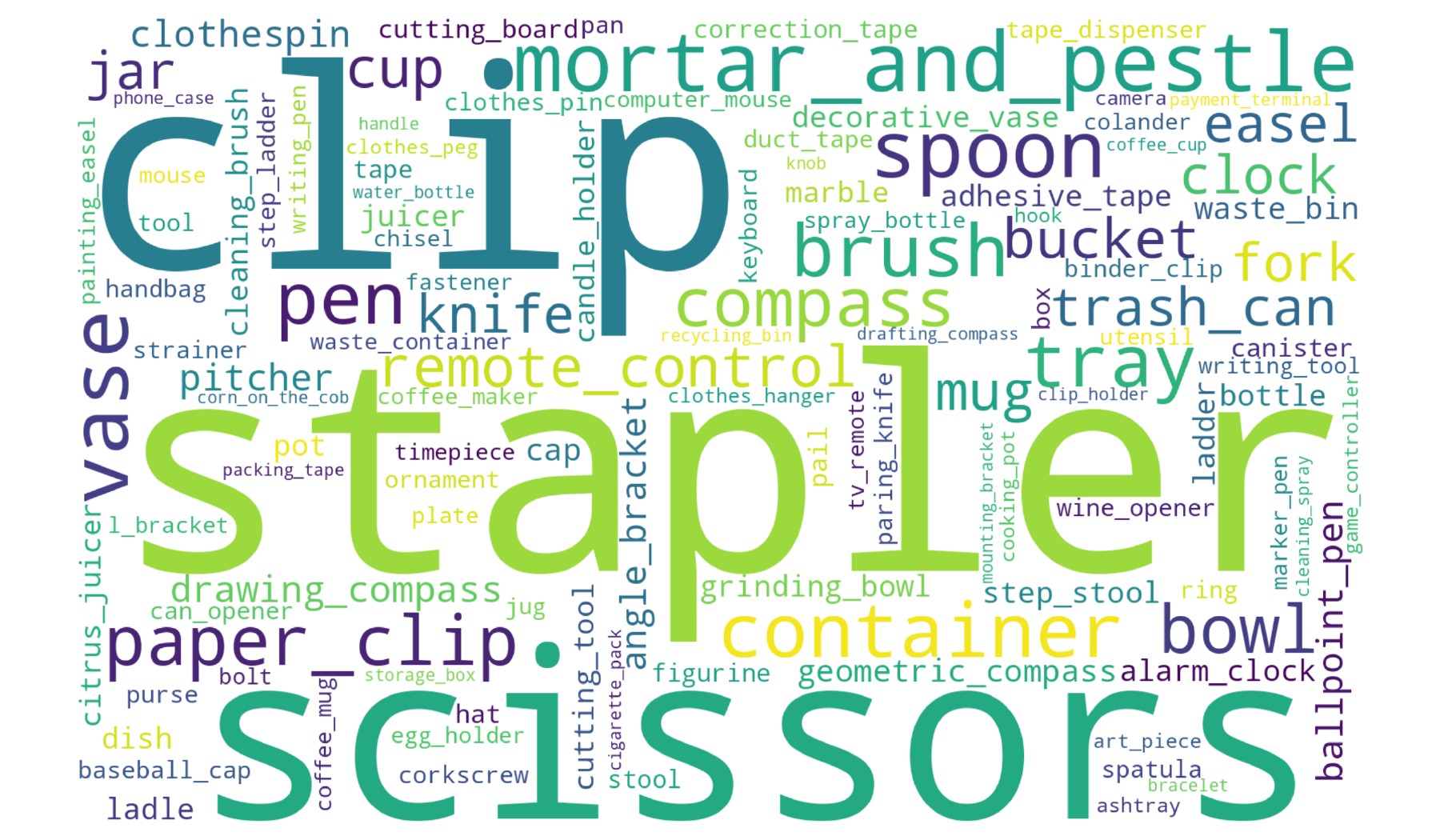}
    \includegraphics[width=0.63\linewidth]{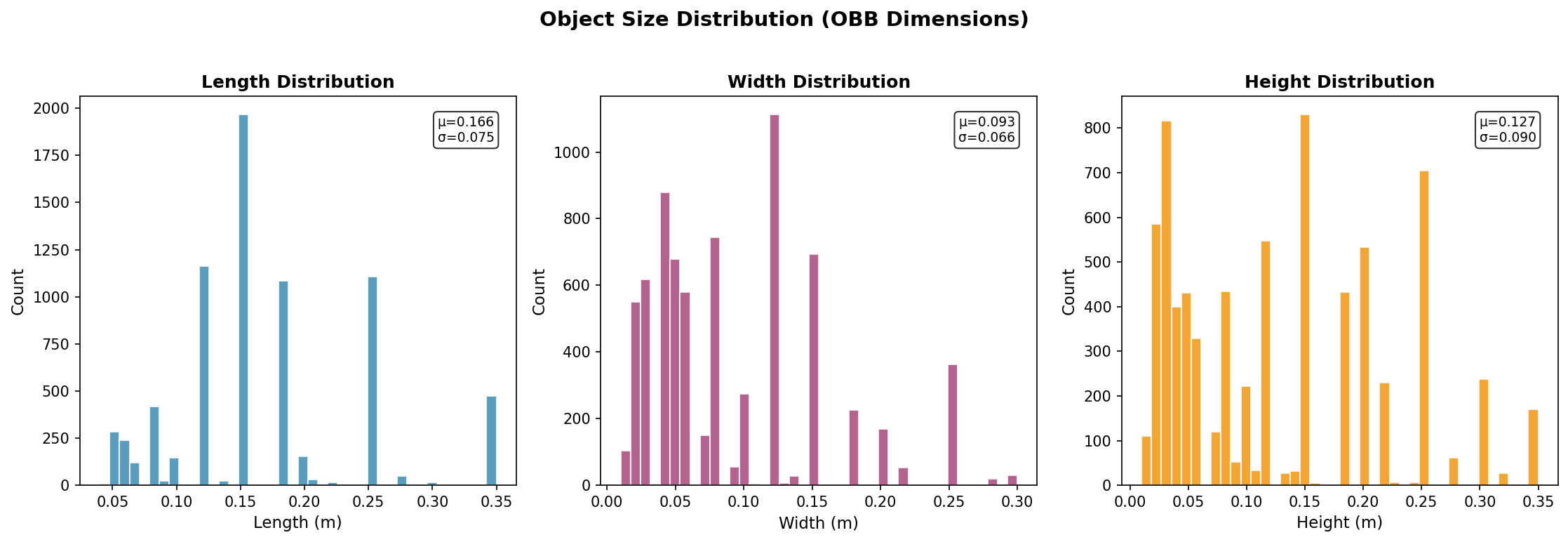}
    \caption{\textbf{Dataset Distribution.} (Left) Word cloud showing object category diversity across ManiTwin-100K. (Right) Distribution of object dimensions, spanning typical manipulation scales from 5--50\,cm.}
    \label{fig:size_distr}
\end{figure}

\begin{figure}[t]
    \centering
    \includegraphics[width=0.195\linewidth]{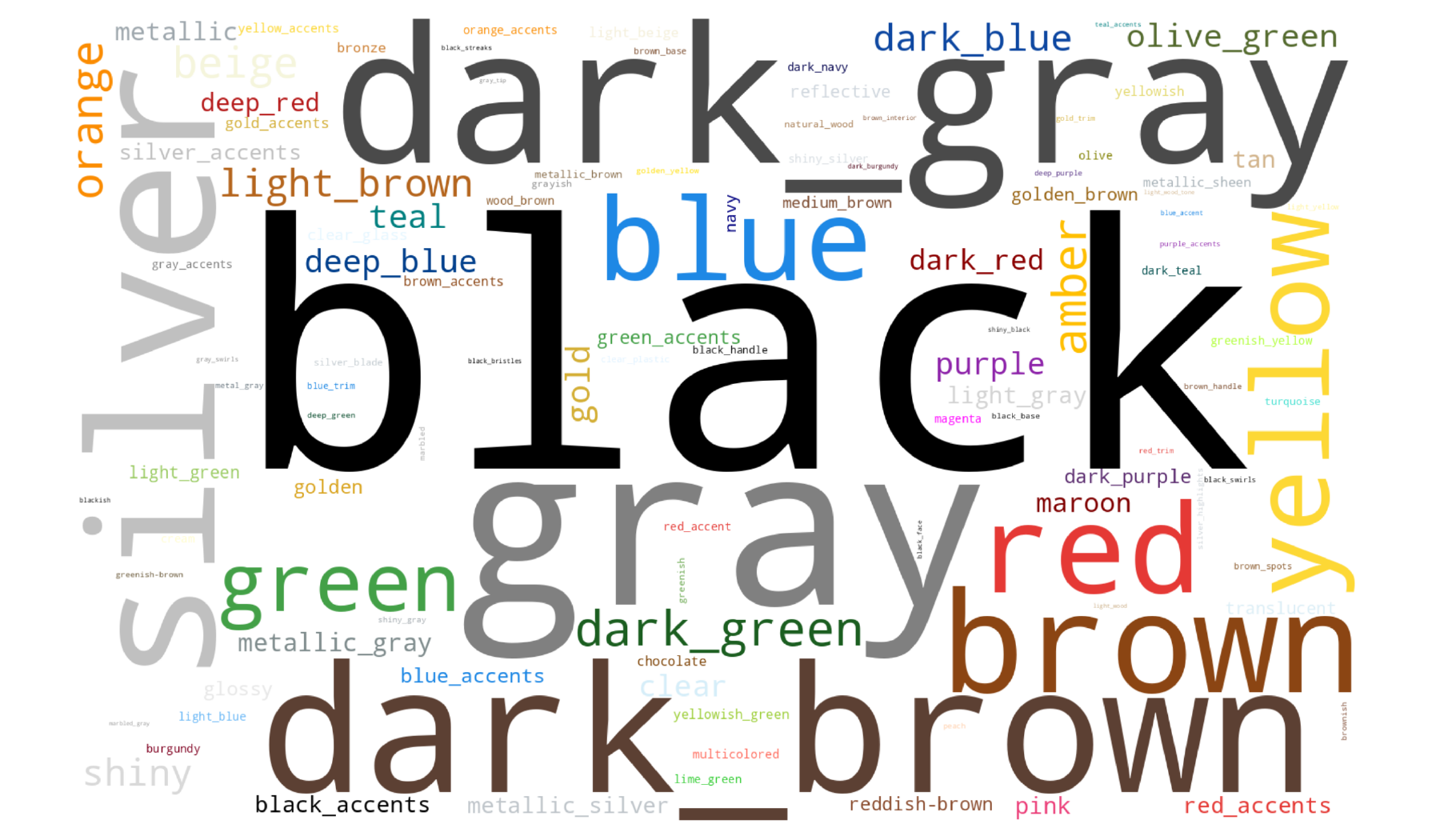}
    \includegraphics[width=0.195\linewidth]{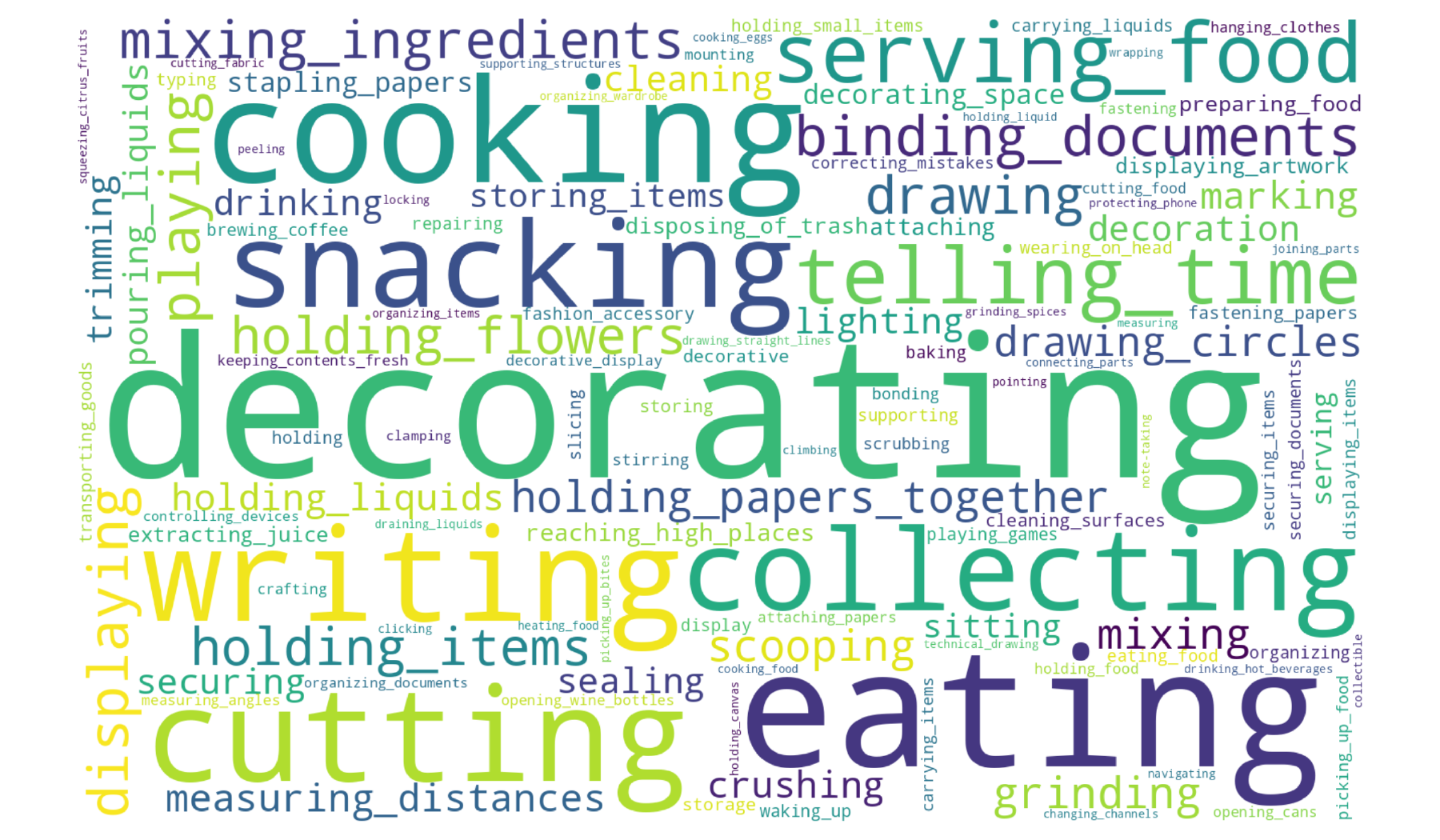}
    \includegraphics[width=0.195\linewidth]{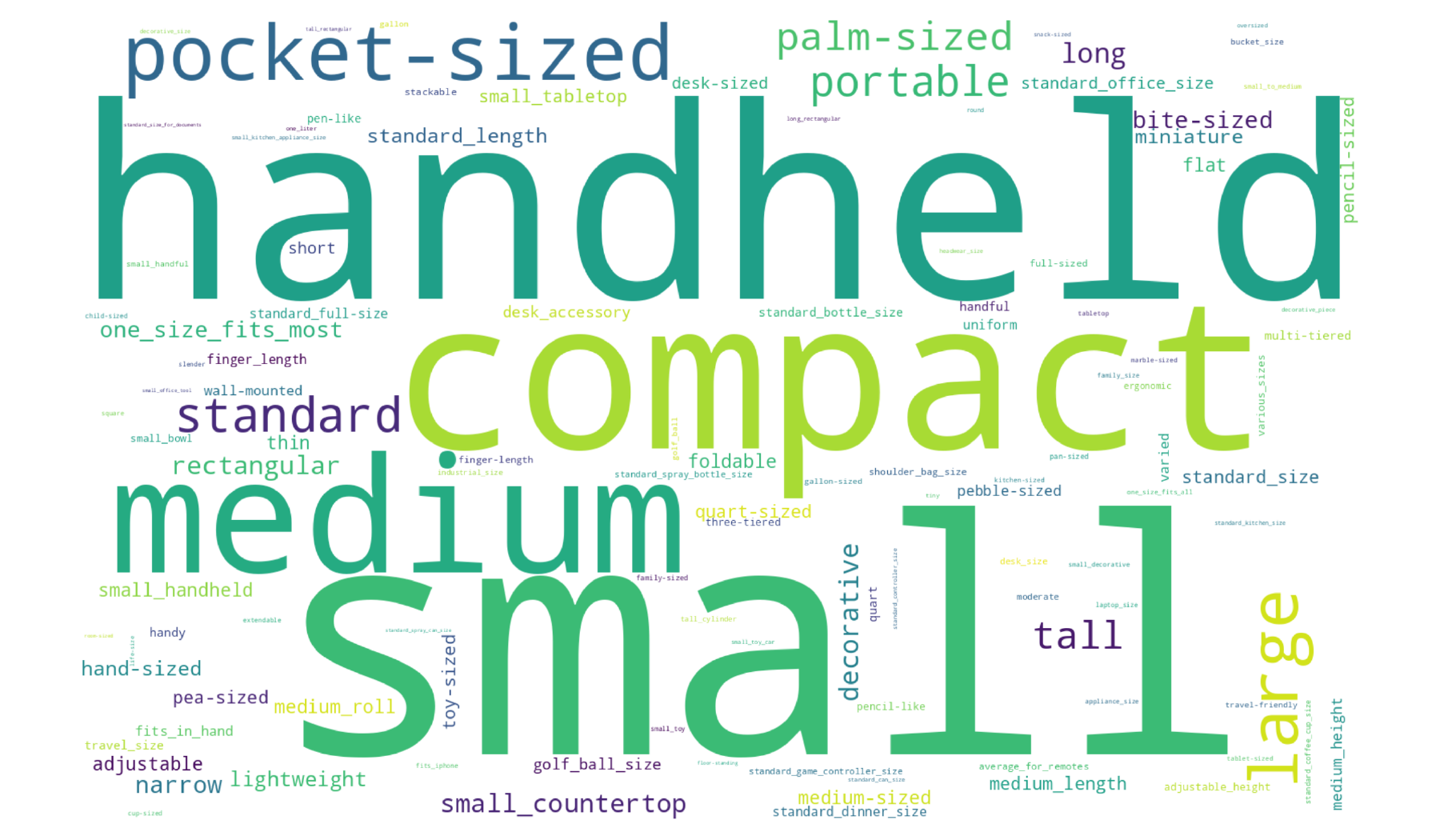}
    \includegraphics[width=0.195\linewidth]{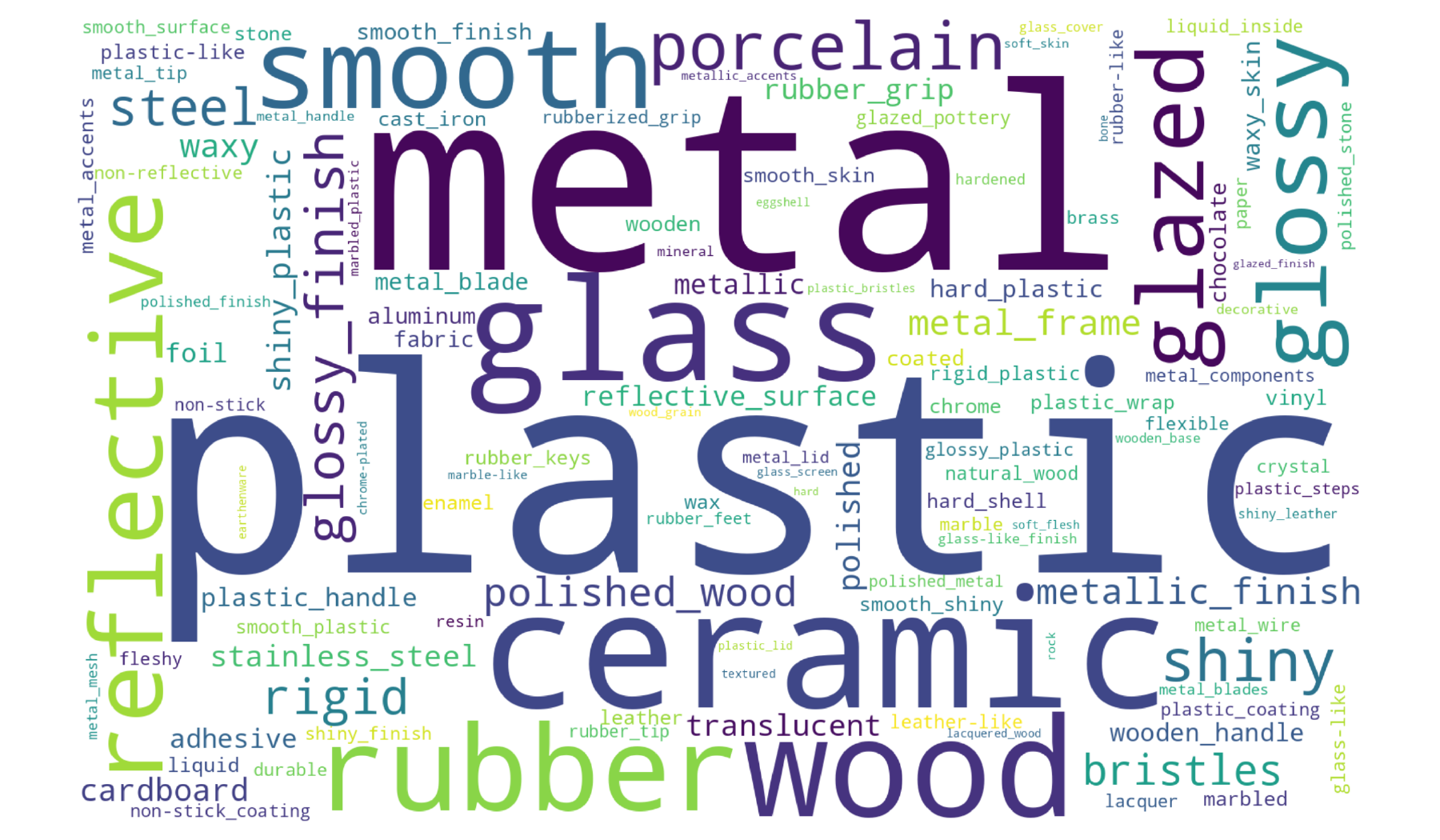}
    \includegraphics[width=0.195\linewidth]{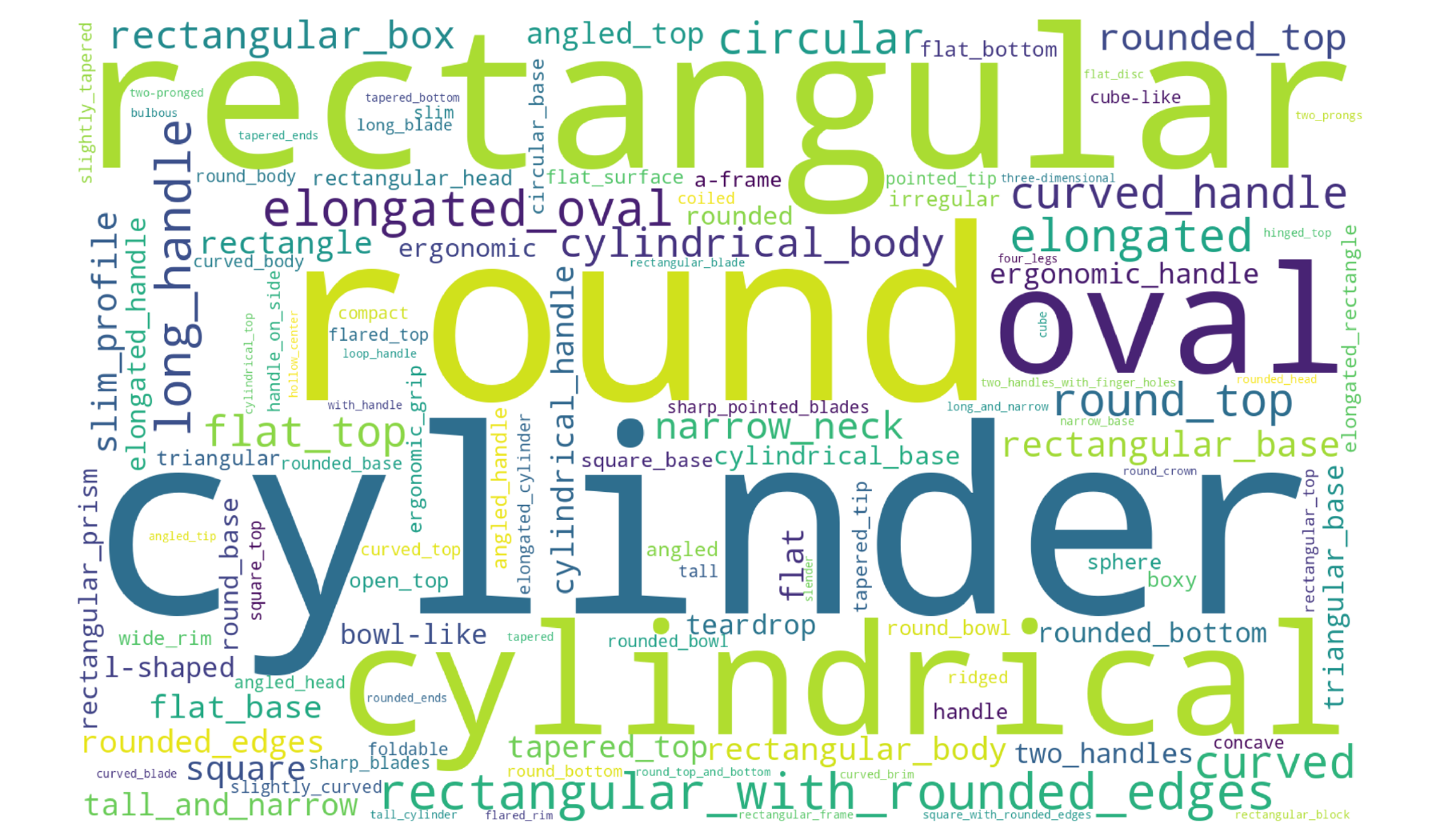}
    \caption{\textbf{Semantic Diversity.} Word clouds for object attributes: color, function, size, category, material, and shape.}
    \label{fig:word_cloud}
\end{figure}

We analyse the semantic and object diversity of ManiTwin-100K dataset. The dataset covers 512 object categories. The semantic diversity is illustrated in Fig.~\ref{fig:word_cloud}, with category distribution and size statistics shown in Fig.~\ref{fig:size_distr}. Object dimensions span from 2\,cm (small tools, cosmetics) to 37\,cm (large containers, tools), covering the typical range encountered in household and industrial manipulation scenarios.

The results demonstrate highly diverse object categories and semantic labels, which can benefit diverse task and manipulation data generation. 

\subsection{Data Generation Statistics}

To demonstrate the scalability of ManiTwin for downstream applications, we report statistics on large-scale grasping data generation. Using the simulation-verified grasp poses and placement annotations in ManiTwin, we automatically generate cross-embodiment manipulation trajectories across the full dataset. Table~\ref{tab:datagen-stats} summarizes the data generation statistics.

The dataset contains 100K objects with over 5 million simulation-verified grasp poses. From these, we generate 10 million grasp trajectories with an average length of 7.6 seconds. This scale of manipulation data, automatically generated without human teleoperation, demonstrates the practical utility of ManiTwin for large-scale robotic learning. The combination of diverse objects, verified grasps, and functional annotations enables training manipulation policies that generalize across object categories and task types.

%% file: section/conclusion.tex
\section{Conclusion}

We present ManiTwin, an automated pipeline for generating data-generation-ready digital object twins, and ManiTwin-100K, a large-scale dataset of 100K objects constructed using this pipeline. ManiTwin transforms single input images into simulation-ready 3D assets with physical properties, functional point annotations, grasp configurations, and language descriptions, all validated through physics-based simulation.

The key contribution lies in unifying scale, semantic richness, and physical usability. Unlike geometry-focused datasets requiring extensive manual curation, ManiTwin-100K assets are directly deployable in physics simulators with collision-ready meshes and verified manipulation annotations. We demonstrated utility across multiple applications: cross-embodiment manipulation data generation, scene layout synthesis, robotics VQA curation, and 3D understanding tasks. Experiments validate annotation quality with over 90\% human-evaluated accuracy.

By providing manipulation-centric assets at unprecedented scale, ManiTwin-100K establishes a foundation for training generalizable manipulation policies in simulation. The automated nature of the ManiTwin pipeline also enables continuous expansion of the dataset as 3D generation technology improves, supporting the growing demands of robotic learning systems.

\textbf{Limitations.}
ManiTwin-100K currently covers rigid graspable objects but excludes articulated objects (drawers, doors) and deformable objects (cloth, rope). Physical property estimates are VLM-inferred rather than ground-truth calibrated. Extending the pipeline to articulated structures and incorporating real-world calibration are directions for future work.

%% file: section/appendix.tex
\section{Appendix}

\subsection{Dataset Examples}
\label{sec:supp_examples}

\begin{figure*}[h]
    \centering
    \includegraphics[width=\linewidth]{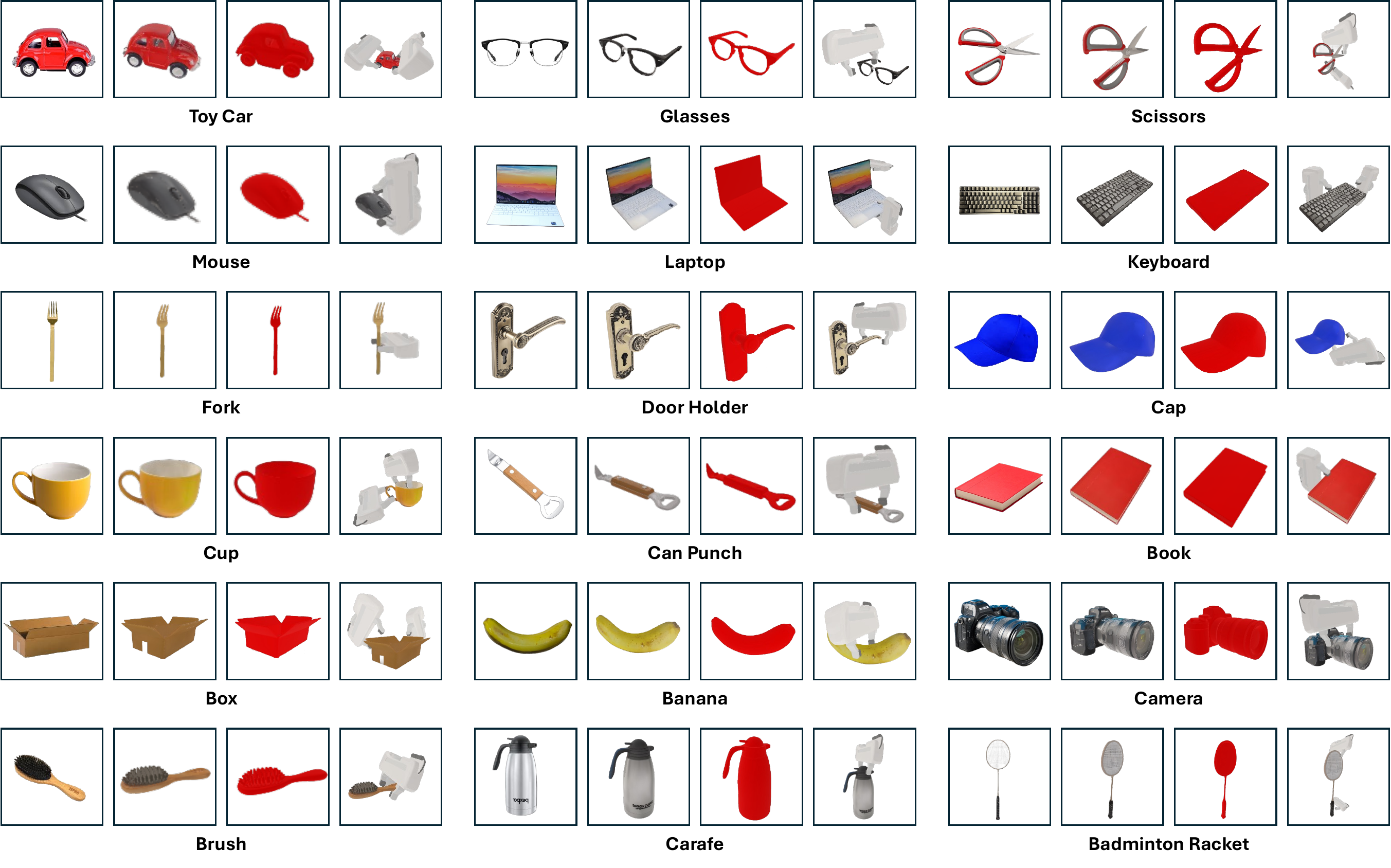}
    \caption{\textbf{ManiTwin-100K Dataset Examples.} Each row shows one object. From left to right: input image, generated 3D asset, mesh visualization, and samples of simulation-verified grasp poses.}
    \label{fig:dataset_examples}
\end{figure*}

Figure~\ref{fig:dataset_examples} presents representative examples from ManiTwin-100K, illustrating the full pipeline from input image to annotated digital twin. Each row shows a single object with four visualization stages:

\textbf{Input Image.} The leftmost column shows the input image used for 3D generation. These images are sourced from diverse origins including e-commerce product photos, rendered views from existing 3D repositories, and text-to-image generations.

\textbf{Generated 3D Asset.} The second column displays the 3D asset produced by our generation pipeline, rendered from a canonical viewpoint. The generated meshes faithfully preserve the geometric structure and visual appearance of the input images, including fine details such as handles, spouts, buttons, and surface textures.

\textbf{Mesh Visualization.} The third column shows the underlying mesh geometry, revealing the mesh topology and density. Our pipeline produces clean, watertight meshes suitable for physics simulation, with appropriate polygon density to capture geometric details while remaining computationally efficient for collision detection.

\textbf{Sampled Grasps.} The rightmost column visualizes a subset of simulation-verified grasp poses. Grasp poses are shown as gripper visualizations positioned at the predicted 6-DoF configurations, representing diverse approach directions and grasp locations validated through physics simulation.

\subsection{Pipeline Implementation Details}
\label{sec:supp_implementation}

\subsubsection{3D Generation}

We use CLAY\cite{zhang2024clay} as the primary image-to-3D generation model. For each input image, we generate a 3D mesh in USDZ format, which is then converted to GLB for downstream processing. Generation takes approximately 45 seconds per object. Failed generations (non-watertight meshes, degenerate geometry) are automatically filtered, resulting in a 69.67\% success rate.

\subsubsection{Physical Property Estimation}

The VLM analyzes 8 multi-view renders of each object to estimate:
\begin{itemize}
    \item \textbf{OBB Dimensions}: Length, width, and height in meters
    \item \textbf{Mass}: Estimated in kilograms based on apparent material and size
    \item \textbf{Friction Coefficient}: Based on surface material (e.g., 0.3 for plastic, 0.5 for rubber)
\end{itemize}
Objects are rescaled to match VLM-estimated real-world dimensions using the longest OBB axis as reference.

\subsubsection{Point Sampling and Selection}

We sample 20,000 points uniformly from the mesh surface, then apply Farthest Point Sampling (FPS) to select 42 candidate points that maximize spatial coverage. The VLM evaluates each candidate against multi-view renders to identify functional regions and suitable grasp locations.

\subsubsection{Grasp Generation and Filtering}

GraspGen produces up to 4,000 grasp candidates per object using the Franka Panda gripper model. We filter grasps by: (1) proximity to VLM-selected grasp points within 3cm threshold; (2) 7-DoF FPS for diversity, retaining 100 representative grasps; (3) simulation verification in SAPIEN\cite{xiang2020sapien}.

\subsubsection{Simulation Verification}

Each candidate grasp undergoes physics simulation with SAPIEN (PhysX 5.0 backend), maximum 2,000 simulation steps, requiring 3 consecutive stable frames with less than 0.01m displacement. A grasp passes if the object remains stably grasped without collision penetration throughout the lift trajectory.

\subsection{Failure Cases and Filtering}
\label{sec:supp_failures}

Our multi-stage pipeline incorporates quality filtering at each stage:

\textbf{3D Generation Failures.} Objects may be rejected due to incomplete geometry where parts are missing or hollow, incorrect topology with self-intersecting meshes, or texture artifacts.

\textbf{VLM Quality Verification Failures.} The VLM-based quality checker rejects assets containing multiple objects instead of a single coherent item, or assets that appear broken, melted, or visually corrupted.

\textbf{Grasp Verification Failures.} Grasps fail simulation verification due to collision between gripper and object, unstable grasp where object slips during lift, or unreachable gripper configurations.
